\documentclass[11pt]{article}
\usepackage{SOTA_style}

\usepackage{lipsum}  

\title{State of the Art: Face Recognition
}
\author{Rubel Biswas\\
    \href{mailto:rbis@unileon.es}{\texttt{rubel.biswas@unileon.es}} 
    \and Pablo Blanco-Medina\\
    \href{mailto:pblanm@unileon.es}{\texttt{pablo.blanco@unileon.es}} 
    }
\date{}    

\begin{document}
{\setstretch{.8}
\maketitle
\begin{abstract}

Working with Child Sexual Exploitation Material (CSEM) in forensic applications might be benefited from the progress in automatic face recognition. However, discriminative parts of a face in CSEM, i.e., mostly the eyes, could be often occluded to difficult the victim's identification. Most of the face recognition approaches cannot deal with such kind of occlusions, resulting in inaccurate face recognition results. This document presents a short review face recognition methods for images with natural and eye occlude faces. The purpose is to select the best baseline approach for solving automatic face recognition of occluded faces.

\noindent

\end{abstract}
}


\section{Introduction}
Face recognition is one of the most broadly researched topics in academic and industrial fields, due to its extensive applications in law enforcement and surveillance, information security, access control, smart cards, and others. In recent years, Deep Convolutional Neural Networks (CNNs) based methods have improved performance significantly \cite{Deng2019ArcFace,Iranmanesh2020GANFaceRec, amos2016openface, Massoli2020CrossFaceRec}.


\par

Fast detection of Child Sexual Exploitation Material (CSEM) could prevent its distribution as soon as possible and would allow building a legal case against presumed offenders.  However, manual detection of such material is time-consuming and disturbing for LEA operators.   

In the context of forensic tools, the automatic detection of CSEM \cite{chaves2020improving} represents a substantial assistance to Law Enforcement Agencies (LEA). In these tools, face recognition can support the task of victim identification in CSEM, apart from establishing links between different CSEM cases\cite{Gangwar2017Pornographydetection}.

\begin{figure}[h]
\begin{center}
\includegraphics[width=0.5\linewidth]{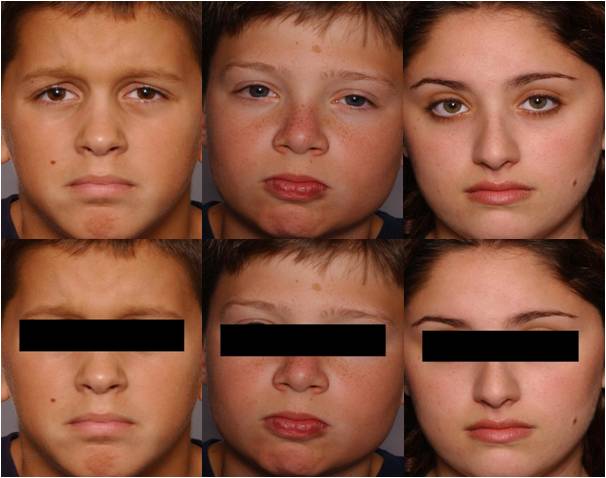}
\end{center}
\caption{Examples of eye occluded versions (bottom) of some faces (top).}
\label{fig:occludedFaceExam}
\end{figure}

However, in CSEM, it is frequent that the offenders manually occlude the victim faces to difficult their identification, e.g. Figure \ref{fig:occludedFaceExam}, which will pose a challenge to face recognition algorithms \cite{biswas2019Boosting}. Under these occluded conditions, the performance of face recognition algorithms drops, mainly when a mask covers the eyes, an object, or by an adversarial attack \cite{biswas2019Boosting}., which is an image modification whose intention is to alter or perturb an image to fool a classifier \cite{Vakhshiteh2020ThreatOA}.

Some approaches to handle the occlusion challenge in the face recognition task have been presented in the last few years. There are frameworks based on auto-encoders \cite{mathai2019does, gorgel2019face}, that remove the occluded parts of the face prior to the recognition. Other approaches have proposed local feature learning-based methods, such as constraints-based dictionary learning \cite{dong2019occlusion}.  Lastly, some approaches have focused on the unbalance between negative and positive samples, such as sparse coding with manifold learning \cite{zhang2016sparse}, or kernel prototype similarities \cite{klare2013heterogeneous}. 

One of the main limitations of this kind of approach is that the augmented examples are extremely correlated to the original ones. Furthermore, most of the methods require training data of both natural and occluded faces of an individual to carry out face recognition against occlusion.



\section{Face Recognition}
Nowadays, significant development has been achieved in face recognition research \cite{Deng2019ArcFace, amos2016openface,Schroff2015}, mainly due to the availability of massive data and Graphical Processing Unit (GPU)s to train Deep Learning models. Some existing approaches \cite{bharadwaj2016domain, zheng2015triangular} have also focused on face recognition with expression, pose, aging, disguise, or illumination changes.

\par
\subsection{Traditional Face Recognition Methods}
Several traditional face recognition methods have been found in the literature, and one of the first approaches has been proposed by Ojala et al. \cite{Ojala1996feature} using  Extended LBP (ELBP) to improve the discriminative capability of faces. The ELBP generates a binary comparison between the central pixel and its neighbors as well as encodes their exact grey-value differences using some additional binary units. 

Wang et al. introduced stationary wavelet entropy to extract features and employed a single hidden layer feedforward neural network as a face classifier \cite{Wang2018Jayaalgo}. Moreover, they also introduced the Jaya algorithm to prevent the training of the classifier fall into local optimum points. 

Model and geometry-based methods are other strategies to recognize a face in unconstrained scenarios. Yin et al. proposed a new model, named Associate Predict (AP) Model \cite{Yin2011assocFR}, to handle the similarity between human faces under significantly different pose, illumination, and expression settings in face recognition. Yang et al. proposed a discriminative Multi-Dimensional Scaling (MDS) method to learn a mapping matrix, which projects the high-resolution and low-resolution face images to a common subspace \cite{Yang2018LowresFace}. They add an inter-class constraint to enlarge the distances of different subjects in the subspace to ensure discriminability.

\subsection{Face Recognition using Deep Learning Methods}

The previous methods are all based on hand-crafted features, and those features may fail to capture important feature information to discriminate against the faces. Nowadays, deep learning-based approaches have become more popular because those methods can deal with large training datasets by learning the rich and compact representation of the face, such as FaceNet \cite{Schroff2015}, VGGFace \cite{simonyan2014very}, and DeepFace \cite{Sun2014DeepID}. Georgescu et al. proposed a hybrid approach that combines the features learned by CNN and hand-crafted features calculated using bag-of-visual-words (BOVW) \cite{Georgescu2019deepHandFeature}. Lastly,  Lu et al. \cite{Lu2018LowRes_FaceRecog} introduced a deep coupled ResNet (DCR) model for low-resolution face recognition. This model consists of two small branch networks and a big trunk network. The trunk and branch networks have been trained to learn discriminant features shared by face images of different resolutions, and to learn resolution-specific coupled-mappings (CMs), respectfully. They projected the high-resolution gallery images and low-resolution images to space, where their distances were minimized, to recognize the face.


\subsection{Face recognition against Occlusion}

As previously mentioned, occlusion is considered one of the most challenging problems in face recognition. Both deep CNN-based or traditional face recognition methods cannot function well against occlusion due to large intra-class variation and higher inter-class similarity.

Apart from Deep learning-based methods, face recognition against occlusion has been handled using different approaches. For instance, Morelli Andr\'{e}s et al. \cite{Morelli2014} employed compressed sensing to detect the occluded part from the face image and then remove it. They used local features to generate a new, non-occluded image that is similar to the one they attempted to recognize. A query image was then subtracted from this new image to detect the occlusion area through a threshold. In the end, only the non-occluded pixels were used to recognize the identity. 

Dagnes et al. \cite{dagnes20193dOccludedFaceRecog} proposed a method for 3D face recognition, robust to the eye and mouth occlusions.  These obstructions were detected and removed by exploiting the 3D geometry, i.e., by considering their effects on the 3D points. Lastly, the non-occluded symmetrical regions were used to restore the missing facial information prior to recognizing the face. 

\par

Domingo et al. \cite{Domingo2014} represented the query and the gallery images by means of random patches,   described by their location and intensity information. These patches were used later to build a dictionary. Wu et al. \cite{Wu2016} proposed a method, called Occlusion Pattern-based Sparse Representation Classification (OPSRC), to learn the occlusion pattern from the query data. Mustafa et al. presented an occluded face recognition framework \cite{Mustafa2017} based on the two-dimensional Multi-Color Fusion (2D-MCF) representation and the Partitioned-sparse sensing recognition (P-SRC) classifier.

\par

Local feature learning is another approach to deal with occlusion. In this procedure, features are extracted from local areas of the face image, and they are used for the recognition through a locally matching strategy. Based on this concept, Liao et al. \cite{Shengcai2013} presented the Multi-Keypoint Descriptors (MKD) to represent the alignment-free face where the actual content of the image determines the size of the descriptor. 

To recognize the partially occluded face, Duan et al. \cite{Duan2018} proposed a scheme based on topology-preserving graph matching to estimate more accurate and robust topological information. It has estimated a non-rigid transformation encoding the second-order geometric structure of the graph.

Non-negative matrix factorization (NMF)-based learning provides an effective way for face recognition robust against occlusions. 
An example is the dictionary learning method proposed by Ou et al. \cite{Weihua2018NMFDictionarylearningOccFaceRecog}. They created low-dimensional representations of samples from the same class to be as close as possible to enhance the discriminant ability of the dictionary.

On the contrary, an LSTM-autoencoders model was introduced by Zhao et al. \cite{zhao2018robust} which consists of a multi-scale spatial LSTM encoder to generate an occlusion-robust representation of the face, and a dual-channel LSTM decoder to recurrently remove the occlusion in the image space.

Additionally, several recent approaches have addressed the occlusion problem by employing low-rank representations \cite{Guangwei2017}, hierarchical sparse and low-rank representations, \cite{Ying2018}, a discriminative multi-scale sparse coding (DMSC) \cite{Feng2017} and fuzzy max-pooling to solve the double-occlusion problem \cite{Long2018}.

Table \ref{Tab:FullOcc} shows the face classification experiment with the datasets, Labeled Faces in the Wild (LFW) \cite{Erik2007}, which has $13,233$ images belonging to $5,749$ identities, and Celebrities in Frontal-Profile in the Wild (CFPW) \cite{Sengupta2016}, which has $7,000$ images of $500$ identities. Note that we created eye occluded versions of both datasets by occluding the eye region of their faces. In face classification against occlusion, embeddings of natural faces using CNNs models, i.e., Dlib \cite{Sharma2016}, VGG16 \cite{simonyan2014very}, OpenFace \cite{amos2016openface}, FaceNet \cite{Schroff2015}, and ArcFace \cite{Deng2019ArcFace} were extracted  as well as the face features were computed using the OSF-DNS hashing method [cite it later], which are then used to train a Support Vector Machine (SVM)  classifier. Note that five different SVM kernel functions, i.e., Linear, RBF, Ploy-2,4, and 6, have been used to train the natural facial features, i.e., five individual trained models have been built to evaluate the occluded face classification performance.  Later, CNN embeddings or OSF-DNS hashing features of the face images from the occluded version of the dataset were computed using the aforementioned CNNs models and perceptual hashing method classified by the SVM to retrieve the face identity.

\begin{table}[!ht]
\caption{Face classification against eye occlusion results obtained by SVM with different kernel functions using the descriptors obtained with Dlib \cite{Sharma2016}, VGG16 \cite{simonyan2014very}, OpenFace \cite{amos2016openface}, FaceNet \cite{Schroff2015}, and ArcFace \cite{Deng2019ArcFace}, and with a perceptual hashing method, OSF-DNS [cite it after accepting it in a Journal].}
\label{Tab:FullOcc}
\centering
\footnotesize
\begin{tabular}{c@{\hspace{0.5cm}}c@{\hspace{0.5cm}}c@{\hspace{0.25cm}}c@{\hspace{0.25cm}}c@{\hspace{0.25cm}}c@{\hspace{0.25cm}}c@{\hspace{0.25cm}}c@{\hspace{0.25cm}}c}
\hline
\textbf{Dataset} & \textbf{Kernels} & \textbf{Dlib} & \textbf{VGG16} & \textbf{OpenFace} & \textbf{FaceNet} & \textbf{ArcFace} & \textbf{OSF-DNS} \\
\hline
\multirow{5}{*}{LFW} & Linear & 4.00 & 12.67 & 5.30 & 13.45 & 6.58 & \textbf{82.74} \\
 & RBF & 9.36 & 4.81 & 11.20 & 4.53 & 4.01 & \textbf{45.34} \\
 & Poly-2 & 19.25 & 4.81 & 21.50 & 14.23 & 6.82 & \textbf{89.53} \\
 & Poly-4 & 27.39 & 4.89 & 9.50 & 20.02 & 4.33 & \textbf{88.14} \\
 &  Poly-6 &  23.56 & 4.54 & 5.30 & 20.02 & 2.78 & \textbf{86.06} \\
\hline 
\hline
\multirow{5}{*}{CFPW}  & Linear &\textbf{48.78} & 14.56 & 37.42 & 51.94 & 3.23 &  26.00 \\
 &  RBF &  50.94 & 7.82 & 5.00 & \textbf{54.14} & 2.00 &  18.18 \\
 &  Poly-2 &  46.82 & 6.82 & 29.38 & \textbf{49.26} & 3.75 & 38.52 \\
 &  Poly-4 &  31.02 & 3.01 & 10.56 & 22.06 & 2.33 & \textbf{58.60} \\
 &  Poly-6 &  17.28 & 1.29 & 3.92 & 6.68 & 2.05 & \textbf{63.24} \\
\hline 

\end{tabular}
\end{table}

\section{Discussion and Conclusions}

In this document, we have presented a study of the state-of-the-art traditional and deep learning-based methods for recognizing a face from natural images. Then, we have presented a revision of the state-of-the-art face recognition systems against occlusion. 
    
We evaluated five deep feature-based approaches, i.e., Dlib, VGG16, OpenFace, FaceNet, and ArcFace, and a perceptual hashing method, i.e., OSF-DNS, in face classification against occlusion.  

In the case of eye occluded version of the LFW dataset, we observed that OSF-DNS obtains an accuracy of $82.74\%$, $45.34\%$, $89.53\%$, $88.14\%$, and $86.06\%$ using the Linear, RBF, Poly-2, Poly-4, and Poly-6 kernel functions, respectively, which outperformed the results obtained by the features obtained by the rest of the deep learning-based assessed methods.

On the contrary, in the case of the eye occluded version of the CFPW face dataset, Dlib obtained the highest classification accuracy, i.e., 48.78\%, with Linear kernel function, and FaceNet attained the highest classification accuracy using RBF and Poly-2 kernel functions, i.e., 54.14\% and 49,26\% respectively. OSF-DNS performed better for the kernel functions Poly-4 and Poly-6, where it attained accuracies of
58.6\% and 63.24\%, respectively, which are the highest on this dataset.

In addition, the accuracy of OSF-DNS is comparatively lower in the case of the CFPW dataset. The reason may be the CFPW dataset contains faces with different poses and illuminations. Subsequently, features of some faces, e.g., non-frontal faces in CFPW, are similar to each other and, therefore this makes the classification to be inaccurate.

It is observed in Table \ref{Tab:FullOcc} that the classification accuracies of all deep feature-based methods are generally poorer than the perceptual hashing method, OSF-DNS, for both datasets. 
The reason may be that occlusion leads to a distortion of the face embeddings obtained by the convolutional base of the networks, which makes discrimination more difficult.
Mainly, ArcFace attains much worse performance than the rest of the CNN-based approaches.
In ArcFace, a more reliable method to increase the feature distances is applied \cite{Deng2019ArcFace}: an arc-cosine function is applied in the angular domain so that the decision boundaries between features corresponding to different classes are more distant from each other.  During the experiment, we mainly extracted the embedding features of a face and its eye occluded version through the pre-trained ArcFace model. Though both faces are the same except the occluded region (i.e. they belong to the same class), the method provides very different embeddings for them. The reason may be that in ArcFace, the embeddings of the faces are distributed around each feature center toward the hyper-sphere and it uses an additive angular margin penalty between feature and ground truth weight to concurrently enhance the intra-class compactness and inter-class discrepancy. 

In this work, however, the pre-trained model is used, which is trained with non-occluded faces, to extract the embeddings of an occluded face. Therefore, occluded face features for the same identity are distorted, thus affecting the accuracy.

Lastly, the revision of this document concludes that traditional deep learning-based approaches may not perform well in the task of face classification or recognition against occlusion. Besides, the experimental results demonstrate that the OSF-DNS features achieve the highest accuracy with almost all the kernel functions compare to the features obtained from five deep learning techniques:  Dlib,  VGG16,  OpenFace,  FaceNet,  and  ArcFace over two labeled datasets,  i.e.,  LFW  and  CFP. Subsequently, OSF-DNS can be recommended for forensic tools to classify whether an eye occluded face is found in a database of non-occluded faces. This may be helpful to make a legal case involving CSEM or other criminal materials.

\section*{Acknowledgment}

This work was supported by the framework agreement between the University of Le\'on and INCIBE (Spanish National Cybersecurity Institute) under Addendum 01.

\medskip

\bibliography{references.bib} 

\begin{thebibliography}{10}

\bibitem{Deng2019ArcFace}
Deng Jiankang, Guo Jia, Xue Niannan, and Zafeiriou Stefanos.
\newblock Arcface: Additive angular margin loss for deep face recognition.
\newblock In {\em Computer Vision and Pattern Recognition (CVPR)}, pages
  4690--4699, 2019.

\bibitem{Iranmanesh2020GANFaceRec}
Seyed~Mehdi Iranmanesh, Benjamin Riggan, Shuowen Hu, and Nasser~M. Nasrabadi.
\newblock {Coupled generative adversarial network for heterogeneous face
  recognition}.
\newblock {\em Image and Vision Computing}, 94:1–--10, 2020.

\bibitem{amos2016openface}
Brandon Amos, Bartosz Ludwiczuk, and Mahadev Satyanarayanan.
\newblock Openface: A general-purpose face recognition library with mobile
  applications.
\newblock {\em CMU School of Computer Science, (2016)}, 2016.

\bibitem{Massoli2020CrossFaceRec}
Fabio~Valerio Massoli, Giuseppe Amato, and Fabrizio Falchi.
\newblock {Cross-resolution learning for Face Recognition}.
\newblock {\em Image and Vision Computing}, 99:1–--15, 2020.

\bibitem{chaves2020improving}
D.~Chaves, E.~Fidalgo, E.~Alegre, F.~Jánez-Martino, and R.~Biswas.
\newblock Improving age estimation in minors and young adults with occluded
  faces to fight against child sexual exploitation.
\newblock In {\em {VISIGRAPP}}, pages 721--729, 2020.

\bibitem{Gangwar2017Pornographydetection}
Abhishek Gangwar, E~Fidalgo, E~Alegre, and V~Gonz{\'a}lez-Castro.
\newblock Pornography and child sexual abuse detection in image and video: A
  comparative evaluation.
\newblock In {\em 8th International Conference on Imaging for Crime Detection
  and Prevention (ICDP 2017)}, pages 37--42, 2017.

\bibitem{biswas2019Boosting}
Rubel Biswas, V\'ictor Gonz\'alez-Castro, Eduardo Fidalgo~Fern\'andez, and
  Deisy Chaves.
\newblock Boosting child abuse victim identification in forensic tools with
  hashing techniques.
\newblock In {\em Jornadas Nacionales de Investigaci\'on en Ciberseguridad
  (JNIC)}, pages 1--2, 2019.

\bibitem{Vakhshiteh2020ThreatOA}
Fatemeh Vakhshiteh, R.~Ramachandra, and A.~Nickabadi.
\newblock {Threat of Adversarial Attacks on Face Recognition: A Comprehensive
  Survey}.
\newblock {\em ArXiv}, abs/2007.11709, 2020.

\bibitem{mathai2019does}
J.~{Mathai}, I.~{Masi}, and W.~{AbdAlmageed}.
\newblock Does generative face completion help face recognition?
\newblock In {\em 2019 International Conference on Biometrics (ICB)}, pages
  1--8, 2019.

\bibitem{gorgel2019face}
P.~G\"{o}rgel and A.~Simsek.
\newblock {Face recognition via Deep Stacked Denoising Sparse Autoencoders
  (DSDSA)}.
\newblock {\em Applied Mathematics and Computation}, 355:325--342, 2019.

\bibitem{dong2019occlusion}
J.~Dong, L.~Zhang, Y.~Chen, and W~Jiang.
\newblock Occlusion expression recognition based on non-convex low-rank double
  dictionaries and occlusion error model.
\newblock {\em Signal Processing: Image Communication}, 76:81--88, 2019.

\bibitem{zhang2016sparse}
P.~Zhang, X.~You, W.~Ou, C.~P. Chen, and Y.-m. Cheung.
\newblock Sparse discriminative multi-manifold embedding for one-sample face
  identification.
\newblock {\em Pattern Recognition}, 52:249--259, 2016.

\bibitem{klare2013heterogeneous}
B.~F. Klare and A.~K. Jain.
\newblock Heterogeneous face recognition using kernel prototype similarities.
\newblock {\em IEEE transactions on pattern analysis and machine intelligence},
  35:1410--1422, 2013.

\bibitem{Schroff2015}
Florian Schroff, Dmitry Kalenichenko, and James Philbin.
\newblock Facenet: A unified embedding for face recognition and clustering.
\newblock In {\em Proceedings of the IEEE conference on computer vision and
  pattern recognition}, pages 815--823, 2015.

\bibitem{bharadwaj2016domain}
Samarth Bharadwaj, Himanshu~S Bhatt, Mayank Vatsa, and Richa Singh.
\newblock Domain specific learning for newborn face recognition.
\newblock {\em IEEE Transactions on Information Forensics and Security},
  11(7):1630--1641, 2016.

\bibitem{zheng2015triangular}
Lilei Zheng, Khalid Idrissi, Christophe Garcia, Stefan Duffner, and Atilla
  Baskurt.
\newblock Triangular similarity metric learning for face verification.
\newblock In {\em 2015 11th IEEE International Conference and Workshops on
  Automatic Face and Gesture Recognition (FG)}, volume~1, pages 1--7. IEEE,
  2015.

\bibitem{Ojala1996feature}
T.~Ojala, M.~Pietikinen, and D.~Harwood.
\newblock A comparative study of texture measures with classification based on
  featured distributions.
\newblock {\em Pattern Recognition}, 29:51--59, 1996.

\bibitem{Wang2018Jayaalgo}
S.~H. Wang, P.~Phillips, Z.~C. Dong, and Y.~D. Zhang.
\newblock Intelligent facial emotion recognition based on stationary wavelet
  entropy and jaya algorithm.
\newblock {\em Neurocomputing}, 272:668–--676, 2018.

\bibitem{Yin2011assocFR}
Q.~Yin, X.~Tang, and J.~Sun.
\newblock An associate-predict model for face recognition.
\newblock In {\em IEEE Computer Society Conference on Computer Vision and
  Pattern Recognition}, pages 497–--504, 2011.

\bibitem{Yang2018LowresFace}
F.~Yang, W.~Yang, R.~Gao, and Q.~Liao.
\newblock Discriminative multidimensional scaling for low-resolution face
  recognition.
\newblock {\em IEEE Signal Process Letter}, 25(3):388–--392, 2018.

\bibitem{simonyan2014very}
K.~Simonyan and A.~Zisserman.
\newblock Very deep convolutional networks for large-scale image recognition.
\newblock In {\em International Conference on Learning Representations}, pages
  1--14, 2015.

\bibitem{Sun2014DeepID}
Y.~Sun, X.~Wang, and X.~Tang.
\newblock {Deep learning face representation from predicting 10,000 classes}.
\newblock In {\em {In IEEE computer society conference on computer vision and
  pattern recognition}}, pages 1891–--1898, 2014.

\bibitem{Georgescu2019deepHandFeature}
MI. Georgescu, RT. Ionescu, and M.~Popescu.
\newblock Local learning with deep and handcrafted features for facial
  expression recognition.
\newblock {\em IEEE Access}, 7:64827–--64836, 2019.

\bibitem{Lu2018LowRes_FaceRecog}
Za~Lu, X.~Jiang, and A.~Kot.
\newblock Deep coupled resnet for low-resolution face recognition.
\newblock {\em IEEE Signal Processing Letter}, 25(4):526–530, 2018.

\bibitem{Morelli2014}
A.~Morelli Andr\'es, S.~Padovani, M.~Tepper, and J.~Jacobo-Berlles.
\newblock Face recognition on partially occluded images using compressed
  sensing.
\newblock {\em Pattern Recognition Letter}, 36:235--242, 2014.

\bibitem{dagnes20193dOccludedFaceRecog}
N.~Dagnes, F.~Marcolin, F.~Nonis, S.~Tornincasa, and E.~Vezzetti.
\newblock 3{D} geometry-based face recognition in presence of eye and mouth
  occlusions.
\newblock {\em International Journal on Interactive Design and Manufacturing},
  2019.

\bibitem{Domingo2014}
Mery Domingo and Bowyer Kevin.
\newblock Face recognition via adaptive sparse representations of random
  patches.
\newblock In {\em IEEE International Workshop on Information Forensics and
  Security}, pages 13--18, 2014.

\bibitem{Wu2016}
Wu~Cho-Ying and Ding Jian-Jiun.
\newblock Occlusion pattern-based dictionary for robust face recognition.
\newblock In {\em Proceedings of the IEEE International Conference Multimedia
  and Expo}, pages 1--6, 2016.

\bibitem{Mustafa2017}
M.~Alrjebi Mustafa, Pathirage Nadith, Liu Wanquan, and Li~Ling.
\newblock {Face recognition against occlusions via colour fusion using 2D-MCF
  model and SRC}.
\newblock {\em Pattern Recognition Letters}, 95:14--21, 2017.

\bibitem{Shengcai2013}
Liao Shengcai, K.~Jain Anil, and Z.~Li Stan.
\newblock Partial face recognition: An alignment-free approach.
\newblock {\em IEEE Transactions on Pattern Analysis and Machine Intelligence},
  35(5):1193--1205, 2013.

\bibitem{Duan2018}
Duan Yueqi, Lu~Jiwen, Feng Jianjiang, and Zhou Jie.
\newblock Topology preserving structural matching for automatic partial face
  recognition.
\newblock {\em IEEE Transactions on Information Forensics and Security},
  13(7):1823--1837, 2018.

\bibitem{Weihua2018NMFDictionarylearningOccFaceRecog}
Weihua Ou, Xiao Luan, Jianping Gou, Quan Zhou, Wenjun Xiao, Xiangguang Xiong,
  and Wu~Zeng.
\newblock Robust discriminative nonnegative dictionary learning for occluded
  face recognition.
\newblock {\em Pattern Recognition Letters}, 107:41--49, 2018.

\bibitem{zhao2018robust}
F.~Zhao, J.~Feng, J.~Zhao, W.~Yang, and S.~Yan.
\newblock Robust {LSTM}-autoencoders for face de-occlusion in the wild.
\newblock {\em IEEE Transactions on Image Processing}, 27:778--790, 2018.

\bibitem{Guangwei2017}
Gao Guangwei, Yang Jian, Jing Xiao-Yuan, Shen Fumin, Yangg Wankou, and Yue
  Dong.
\newblock Learning robust and discriminative low-rank representations for face
  recognition with occlusion.
\newblock {\em Pattern Recognition}, 66:129--143, 2017.

\bibitem{Ying2018}
Ying Cho, Jian Wu, and Ding Jiun.
\newblock Occluded face recognition using low-rank regression with generalized
  gradient direction.
\newblock {\em Pattern Recognition}, 80:256--268, 2018.

\bibitem{Feng2017}
Yu~Yu-Feng, Dai Dao-Qing, Ren Chuan-Xian, and Huang Ke-Kun.
\newblock Discriminative multi-scale sparse coding for single-sample face
  recognition with occlusion.
\newblock {\em Pattern Recognition}, 66:302--312, 2017.

\bibitem{Long2018}
Long Yang, Zhu Fan, Shao Ling, and Han Junwei.
\newblock Face recognition with a small occluded training set using spatial and
  statistical pooling.
\newblock {\em Information Sciences}, 430--431:634--644, 2018.

\bibitem{Erik2007}
B.~Huang Gary, Mattar Marwan, Berg Tamara, and Learned-Miller Erik.
\newblock Labeled faces in the wild: A database for studying face recognition
  in unconstrained environments.
\newblock Technical Report 7--49, Univ. Massachusetts, USA, 2007.

\bibitem{Sengupta2016}
S.~Sengupta, J.~Chen, C.~Castillo, V.~M. Patel, R.~Chellappa, and D.~W. Jacobs.
\newblock Frontal to profile face verification in the wild.
\newblock In {\em IEEE Winter Conference on Applications of Computer Vision
  (WACV)}, pages 1--9, 2016.

\bibitem{Sharma2016}
S.~Sharma, Shanmugasundaram Karthikeyan, and Kumar~Ramasamy Sathees.
\newblock {FAREC -- CNN based efficient face recognition technique using Dlib}.
\newblock In {\em International Conference on Advanced Communication Control
  and Computing Technologies (ICACCCT)}, pages 192--195, 2016.

\end{thebibliography}

\newpage




\end{document}